\documentclass[letterpaper,twocolumn,fleqn]{article} 

\usepackage{ist}
\usepackage{amsmath}
\usepackage{booktabs}
\usepackage{graphicx}
\usepackage{array}
\usepackage{float}
\usepackage{caption}
\usepackage{hyperref}
\usepackage{amsmath}
\usepackage{amssymb}

\usepackage{cuted}       
\usepackage{capt-of}     
\usepackage{booktabs}    
\usepackage{array}       

\title{Image Segmentation: Inducing graph-based learning}

\author{Aryan Singh$^1$, Pepijn Van de Ven$^1$, Ciarán~Eising$^1$, and Patrick Denny$^1$\\
$^1$Data-Driven Computer Engineering Group, Dept. of Electronic and Computer Engineering, University of Limerick, Ireland}

\date{} 

\begin{document} 

\maketitle 

\thispagestyle{empty} 


\begin{abstract}
This study explores the potential of graph neural networks (GNNs) to enhance semantic segmentation across diverse image modalities. We evaluate the effectiveness of a novel GNN-based U-Net architecture on three distinct datasets: PascalVOC, a standard benchmark for natural image segmentation, WoodScape, a challenging dataset of fisheye images commonly used in autonomous driving, introducing significant geometric distortions; and ISIC2016, a dataset of dermoscopic images for skin lesion segmentation. We compare our proposed UNet-GNN model against established convolutional neural networks (CNNs) based segmentation models, including U-Net and U-Net++, as well as the transformer-based SwinUNet. Unlike these methods, which primarily rely on local convolutional operations or global self-attention, GNNs explicitly model relationships between image regions by constructing and operating on a graph representation of the image features. This approach allows the model to capture long-range dependencies and complex spatial relationships, which we hypothesize will be particularly beneficial for handling geometric distortions present in fisheye imagery and capturing intricate boundaries in medical images. Our analysis demonstrates the versatility of GNNs in addressing diverse segmentation challenges and highlights their potential to improve segmentation accuracy in various applications, including autonomous driving and medical image analysis. \href{https://github.com/aryan-at-ul/Electronic-Imaging-2025-paper-4492.git}{Code Available at GitHub}.
\end{abstract}

\section{INTRODUCTION}
\label{sec:intro}
Image segmentation is a critical component of autonomous driving systems, enabling vehicles to interpret and understand their surroundings by partitioning visual input into meaningful regions. Accurate segmentation allows for the identification of lanes, detection of obstacles, recognition of traffic signs, and comprehension of the dynamic environment necessary for safe navigation. Without precise segmentation, autonomous vehicles would struggle to make informed decisions, potentially compromising safety and efficiency.

Deep Neural Networks (DNNs) have significantly advanced the field of image segmentation, providing powerful tools for feature extraction and pattern recognition. CNNs\cite{cnn}, in particular, have become the backbone of many segmentation models due to their ability to learn hierarchical representations of visual data. Prominent architectures such as U-Net\cite{unet}, U-Net++\cite{unetpp}, and SwinUNet\cite{swinunet} have demonstrated exceptional performance in various segmentation tasks. U-Net and its variants are especially known for their encoder-decoder structures that capture contextual information while preserving spatial details, making them suitable for biomedical and natural image segmentation. SwinUNet integrates the strengths of the Swin Transformer with the U-Net architecture, further enhancing feature representation and capturing long-range dependencies.

Despite their success, CNN-based segmentation methods have inherent limitations. They often struggle to capture the intricate relationships between objects, especially in complex scenes where contextual information is crucial. Additionally, CNNs are sensitive to geometric distortions, such as those introduced by fisheye cameras commonly used in autonomous vehicles to achieve wide fields of view. These distortions can degrade the performance of CNNs, as their convolutional kernels may not effectively handle non-linear transformations in the image space\cite{fisheye}.

Graph-based methods offer a promising solution to these challenges by modeling images as graphs where nodes represent pixels or regions, and edges represent relationships between them. GNNs\cite{gnn} excel at capturing complex dependencies and can naturally handle irregular structures and non-local interactions. By leveraging GNNs, it's possible to consider global contextual information and the geometric relationships between objects\cite{gnnvis}, which is particularly advantageous in scenarios with distorted images or where object relationships play a critical role.

In this work, we propose a hybrid approach that combines CNNs and GNNs to improve image segmentation for autonomous driving. A CNN extracts rich local feature representations, which are used to construct a graph where nodes represent feature vectors, and edges capture spatial and feature similarities. A GNN processes this graph to refine segmentation by modeling complex relationships between image regions. This integration leverages the local pattern extraction of CNNs and the global relational modeling of GNNs, addressing limitations of traditional CNN-based methods \cite{cnnfish}.

We evaluate our proposed method on three distinct datasets to demonstrate its effectiveness and robustness. The first is the PascalVOC Segmentation dataset\cite{pascal}, a widely used benchmark for general-purpose image segmentation that includes a variety of object classes and scene complexities. The dataset contains annotated images of everyday objects, providing a comprehensive testbed for assessing segmentation models. The second is the WoodScape dataset\cite{woodscape}, specifically designed for autonomous driving scenarios with fisheye camera images. WoodScape provides a challenging set of images with significant geometric distortions, multiple camera perspectives, and diverse driving conditions, making it ideal for testing the robustness of segmentation models in real-world driving environments. The third is the ISIC2016 dataset\cite{isic2016}, a medical imaging benchmark for skin lesion segmentation. This dataset contains dermoscopic images with detailed annotations, offering a domain-specific evaluation of segmentation models in the context of healthcare, where precision and sensitivity are critical.

Our experimental results indicate that incorporating GNNs into the segmentation pipeline enhances accuracy, especially in challenging situations with distorted images. On the PascalVOC dataset, our method achieves competitive performance, demonstrating its general applicability. On the WoodScape dataset, we observe a notable improvement over traditional CNN-based methods, highlighting the advantage of our hybrid approach in dealing with geometric distortions. These findings suggest that our CNN-GNN framework can provide more robust and accurate scene understanding, potentially improving the reliability and safety of autonomous driving systems. This work makes two key contributions:

\begin{enumerate}
    \item \textbf{A Novel Approach Combining CNNs and GNNs:} We introduce a new method that harnesses the strengths of both CNNs and GNNs for semantic segmentation. By leveraging CNNs to extract hierarchical features and GNNs to refine them through graph-based interactions, we achieve more accurate and robust segmentation results.
    
    \item \textbf{Validation of Robustness Across Diverse Datasets:} We demonstrate the effectiveness of our approach on three distinct datasets: the versatile PascalVOC Segmentation dataset, the challenging WoodScape dataset featuring fisheye images common in autonomous driving scenarios, and the ISIC2016 dataset, a benchmark for skin lesion segmentation in medical imaging. This showcases the generalizability and robustness of our method across diverse imaging conditions and applications.
\end{enumerate} 
\section{RELATED WORK}

Image segmentation is a fundamental task in autonomous driving systems, enabling vehicles to interpret their surroundings by identifying and classifying objects within a scene. Accurate segmentation informs critical decisions for navigation and safety, such as obstacle avoidance and path planning \cite{safety}.

\subsection{U-Net}

U-Net \cite{unet} is a convolutional neural network architecture designed for precise segmentation, particularly effective for biomedical image segmentation. Its architecture is characterized by an encoder-decoder structure with symmetric skip connections. The encoder path contracts the input image through a series of convolutional layers and max pooling operations, capturing contextual information at different scales. The decoder path then expands the feature maps using up-convolution (transposed convolution) layers, recovering the spatial resolution. Crucially, the skip connections directly transfer high-resolution feature maps from the encoder to the corresponding layers in the decoder. This mechanism helps to preserve fine-grained details and improve localization accuracy, which is essential for identifying small or distant objects on the road in autonomous driving applications \cite{unetseg}. The original U-Net used relatively few training samples, relying on strong data augmentation to achieve robust performance.

\subsection{U-Net++}

U-Net++ \cite{unetpp} builds upon the U-Net architecture by introducing nested and dense skip connections. Instead of direct skip connections between corresponding encoder and decoder levels, U-Net++ uses a series of nested, dense skip connections that connect all encoder and decoder nodes at the same level. This design creates multiple U-Nets of varying depths within the architecture, enabling the aggregation of features from different receptive fields. This nested architecture effectively bridges the semantic gap between encoder and decoder feature maps by gradually refining the features passed through the skip connections. This approach facilitates better feature recalibration, leading to more detailed and accurate segmentation results, especially in complex scenes typical of autonomous driving environments. The deep supervision provided by the multiple U-Nets further enhances the training process. This allows for improved segmentation of objects at multiple scales.

\subsection{SwinUNet}

The Swin Transformer \cite{swint} revolutionized computer vision by effectively applying transformer architectures, originally designed for natural language processing, to image tasks\cite{dosovitskiy2021imageworth16x16words}. SwinUNet \cite{swinunet} integrates the Swin Transformer into the U-Net framework, replacing the convolutional layers in the encoder and decoder with Swin Transformer blocks. The Swin Transformer uses shifted windows for efficient computation of self-attention, allowing the model to capture long-range dependencies and global context while maintaining computational efficiency. This is a significant advantage over traditional CNNs, which have limited receptive fields. In SwinUNet, the hierarchical structure of the Swin Transformer allows for multi-scale feature extraction, similar to the encoder-decoder structure of U-Net. The ability to model complex spatial relationships makes SwinUNet a promising approach for autonomous driving scenarios, where understanding the global context is vital for tasks like detecting distant objects, comprehending road layouts, and predicting the behavior of other road users. Although initially prominent in medical image segmentation, its potential in autonomous driving is increasingly recognized.

\subsection{Challenges with CNN-Based Methods}

Despite their successes, CNN-based models like U-Net, U-Net++, and SwinUNet face challenges in handling geometric distortions and capturing intricate relationships between objects in driving scenes. Fisheye cameras, commonly used in autonomous vehicles for their wide field of view, introduce significant radial and tangential distortions that standard convolutional kernels struggle to process effectively \cite{sphere}. These distortions can lead to inconsistencies in object shapes and sizes, making it difficult for CNNs to accurately segment objects\cite{sphereseg}. Furthermore, while SwinUNet addresses long-range dependencies to some extent, capturing complex relationships between objects, such as their relative positions and interactions, remains a challenge for purely CNN-based or even transformer based approaches without explicit relational modeling. This limitation can lead to decreased segmentation accuracy in distorted images, affecting the vehicle's ability to make safe and reliable decisions.

The following section details our proposed method, which, like U-Net, employs an encoder-decoder structure. However, to enhance feature separability and achieve more precise segmentation, we introduce a GNN based approach at the bottleneck of the architecture. This GNN layer refines the CNN-extracted features by explicitly modeling their relationships, leading to more distinct representations that are then used for upsampling and ultimately, segmentation mask generation.

\begin{figure*}[ht]
    \centering
    \includegraphics[width=\textwidth,height=3in]{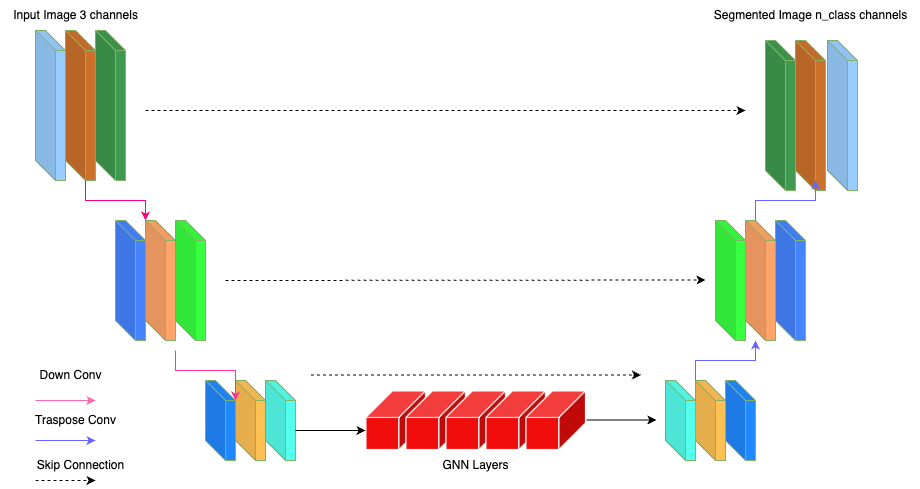}
    \caption{\textbf{Model Architecture.} The network is composed of three main modules: an Encoder, a GNN Bottleneck, and a Decoder.}
    \label{fig:model}
\end{figure*}

\section{Methodology}
\label{sec:methodology}

We propose an enhanced U-Net architecture that incorporates Graph Neural Network (GNN) layers at the bottleneck. This design leverages the convolutional feature extraction capabilities of U-Net while introducing a graph-based mechanism to capture complex global interdependencies—particularly useful for mitigating the geometric distortions introduced by fisheye cameras.

\begin{figure*}[ht]
    \centering
    \includegraphics[width=\textwidth,height=4in]{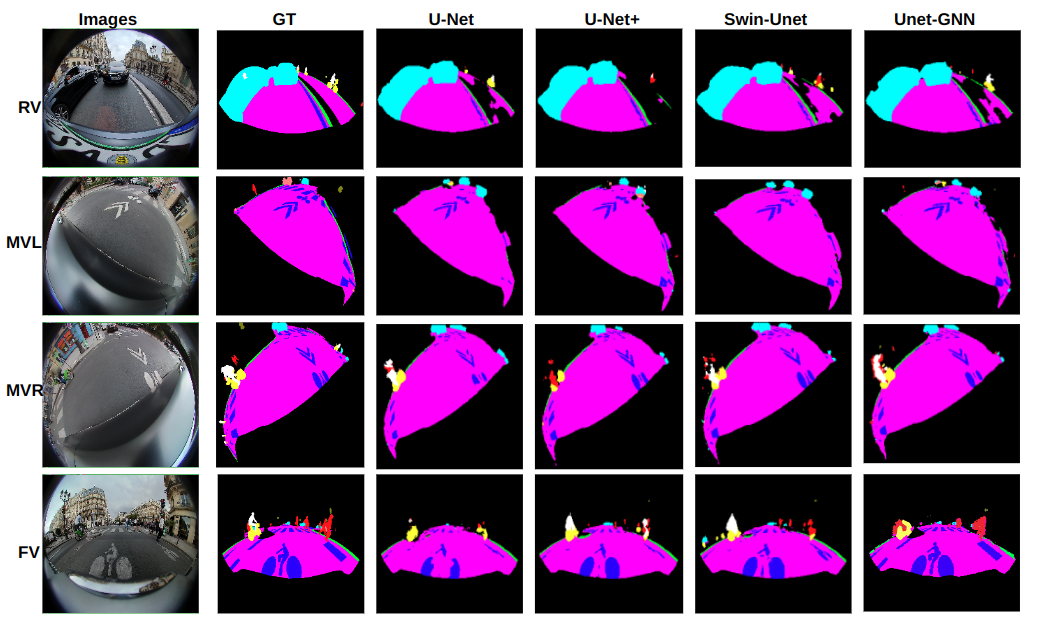}
    \caption{\textbf{Visualization} of segmentation mask generated by different models under this study}
    \label{fig:perf}
\end{figure*}

\subsection{Overall Architecture}

As illustrated in Figure~\ref{fig:model}, the proposed model consists of three primary components:
\begin{itemize}
    \item \textbf{Encoder:} Extracts hierarchical features through convolutional blocks.
    \item \textbf{GNN Bottleneck:} Learns global context and relationships via GNN layers.
    \item \textbf{Decoder:} Reconstructs spatial resolutions and produces the segmentation map.
\end{itemize}


\subsection{Encoder}

Let the input be \( X \in \mathbb{R}^{H \times W \times C} \), all images were resized to dimensions of 256x256. The encoder progressively downsamples \( X \) using a series of convolutional blocks, each consisting of two convolutions, an activation, and a pooling operation:
\begin{equation}
F_l = \text{Pool}\bigl(\sigma(\text{Conv}_2(\sigma(\text{Conv}_1(F_{l-1}))))\bigr),
\label{eq:encoder}
\end{equation}
where:
\begin{itemize}
    \item \(F_l\) is the feature map at block \(l\).
    \item \(\text{Conv}_1\) and \(\text{Conv}_2\) denote convolutional operations (typically with \(3\times3\) kernels).
    \item \(\sigma\) is an activation function (e.g., ReLU).
    \item \(\text{Pool}\) is a pooling operation (e.g., max pooling with a \(2\times2\) kernel and stride~2), halving the spatial dimensions.
    \item \(F_0 = X\).
\end{itemize}

\subsection{GNN Bottleneck}
\label{sec:gnn_bottleneck}

At the bottleneck, we introduce a GNN module to capture long-range dependencies and non-local context, which is particularly beneficial for handling fisheye distortions. Fisheye lenses produce wide fields of view but introduce complex geometric distortions that can adversely affect standard convolution operations \cite{4059340}. Traditional convolutional kernels assume a regular grid structure, making them less robust to non-uniform spatial transformations. In contrast, a graph-based formulation allows for flexible connectivity that can accommodate the irregular geometry of warped images \cite{gcn, Bronstein_2017}.

Specifically, we build a graph from the deepest encoder feature map \( F_L \in \mathbb{R}^{H_L \times W_L \times C_L} \). By connecting each spatial location to its \(k\)-nearest neighbors in a \emph{warped} coordinate space (via relative positional encoding), the GNN bottleneck effectively learns features that are invariant to local distortions. This approach is inspired by the success of graph convolution in capturing relationships on non-Euclidean domains, such as those arising in social networks, point clouds, or other irregular geometries \cite{Bronstein_2017}.

\subsubsection{Graph Construction}

We formulate a graph \(G = (V, E)\) as follows:
\begin{itemize}
    \item \textbf{Nodes} (\(V\)): Each spatial location \((x,y)\) in \(F_L\) corresponds to a node \(v_{xy} \in V\). The initial node feature \(h_{xy}^{(0)} \equiv F_L(x,y)\in \mathbb{R}^{C_L}\). Thus, \(|V| = H_L \times W_L\).
    \item \textbf{Edges} (\(E\)): We create edges based on spatial \(k\)-nearest neighbors (\(k\)-NN). For each node \(v_{xy}\), we connect it to its \(k\) closest nodes in a suitably modified (warped) coordinate space:
    \[
    E = \{(v_{xy},v_{x'y'}) \mid v_{x'y'} \in \mathrm{k\text{-}NN}(v_{xy})\}.
    \]
\end{itemize}

\subsubsection{Relative Positional Encoding and k-NN Search}

To better handle fisheye distortions, we incorporate \emph{relative positional encodings} during the graph construction step. Instead of directly using the spatial coordinates \((x,y)\) when determining neighbors, we add a learned offset \(R_{xy}\) (computed via sine-cosine functions as proposed in \cite{vaswani2023attentionneed}) to warp the coordinates:
\[
P_{xy}' = P_{xy} + R_{xy},
\]
where \(P_{xy}\) is the original spatial coordinate of node \(v_{xy}\) and \(R_{xy}\) is its corresponding relative positional encoding. This warped coordinate space preserves context about local distortions, allowing the GNN to capture relationships in a manner that is more robust to the irregular geometry commonly introduced by fisheye lenses \cite{gnnpos}.

\subsubsection{Graph Convolution}

Once the graph is constructed, node features are updated by aggregating information from neighbors:
\begin{equation}
h_i^{(t)} = \sigma \biggl( \sum_{j \in N(v_i)} W \bigl(h_j^{(t-1)}\bigr) + b \biggr),
\label{eq:gnn_update_bottleneck}
\end{equation}
where:
\begin{itemize}
    \item \(h_i^{(t)}\) is the feature vector of node \(v_i\) at layer~\(t\).
    \item \(W \in \mathbb{R}^{C_L \times C_L}\) and \(b \in \mathbb{R}^{C_L}\) are learnable parameters.
    \item \(\sigma\) is an activation function (e.g., ReLU).
    \item \(N(v_i)\) denotes the neighbors of \(v_i\) (including \(v_i\) itself) under the warped coordinate-based \(k\)-NN.
    \item \(h_j^{(t-1)}\) is the feature vector of neighbor \(v_j\) from the previous layer, with \(h_i^{(0)} = F_L(i)\).
\end{itemize}

By leveraging these graph convolutions, the network aggregates information from spatially non-adjacent but perceptually correlated regions, thus better accounting for the localized stretching and shrinking effects characteristic of fisheye distortions \cite{laplaceg}. This global connectivity significantly improves segmentation accuracy compared to traditional U-Net models, particularly in scenes with large field-of-view imagery.

\subsection{Decoder}

The decoder upsamples features and incorporates skip connections from the corresponding encoder layers:
\begin{equation}
F'_l = \sigma\bigl(\text{ConvTranspose}_l(F'_{l+1}) + S_l\bigr),
\label{eq:decoder}
\end{equation}
where:
\begin{itemize}
    \item \(F'_l\) denotes the feature map at the \(l\)-th decoder block.
    \item \(\text{ConvTranspose}_l\) is a transpose convolution (typically a \(2\times2\) kernel with stride~2).
    \item \(S_l\) is the skip connection feature map from the corresponding encoder block~\(l\).
\end{itemize}


Finally, the segmentation map \( Y \in \mathbb{R}^{H \times W \times C_{\text{out}}} \) is generated by:
\begin{equation}
Y = \text{softmax}\bigl(\text{Conv}(F'_0)\bigr),
\label{eq:segmentation_output}
\end{equation}
where \(C_{\text{out}}\) is the number of output classes.

\begin{table*}[ht]
\centering
\renewcommand{\arraystretch}{0.8} 
\caption{Class-specific accuracy and IoU scores for various models in this study.}
\label{tab:results}
\resizebox{\textwidth}{!}{%
\begin{tabular}{clcccccccc}
\toprule
\textbf{Sr. \#} & \textbf{Categories} 
& \multicolumn{2}{c}{\textbf{U-Net}} 
& \multicolumn{2}{c}{\textbf{U-Net++}} 
& \multicolumn{2}{c}{\textbf{Swin-UNet}} 
& \multicolumn{2}{c}{\textbf{UNet-GNN}} \\
\cmidrule(lr){3-4} \cmidrule(lr){5-6} \cmidrule(lr){7-8} \cmidrule(lr){9-10}
& & \textbf{Acc $\uparrow$} & \textbf{IoU $\uparrow$} 
& \textbf{Acc $\uparrow$} & \textbf{IoU $\uparrow$} 
& \textbf{Acc $\uparrow$} & \textbf{IoU $\uparrow$} 
& \textbf{Acc $\uparrow$} & \textbf{IoU $\uparrow$} \\
\midrule
1  & Background     & 0.99 & 0.94 & 0.99 & 0.98 & 0.99 & 0.96 & 0.99 & 0.97 \\
2  & Road           & 0.98 & 0.90 & 0.98 & 0.94 & 0.96 & 0.92 & 0.97 & 0.94 \\
3  & Lanemark       & 0.73 & 0.64 & 0.48 & 0.44 & 0.60 & 0.56 & 0.78 & 0.69 \\
4  & Curb           & 0.34 & 0.39 & 0.21 & 0.17 & 0.55 & 0.48 & 0.66 & 0.60 \\
5  & Person         & 0.16 & 0.09 & 0.40 & 0.37 & 0.34 & 0.28 & 0.51 & 0.44 \\
6  & Rider          & 0.31 & 0.35 & 0.18 & 0.13 & 0.50 & 0.41 & 0.49 & 0.37 \\
7  & Vehicles       & 0.84 & 0.77 & 0.94 & 0.88 & 0.91 & 0.85 & 0.94 & 0.90 \\
8  & Bicycle        & 0.54 & 0.53 & 0.70 & 0.57 & 0.65 & 0.54 & 0.77 & 0.43 \\
9  & Motorcycle     & 0.28 & 0.23 & 0.21 & 0.14 & 0.54 & 0.42 & 0.74 & 0.60 \\
10 & Traffic Sign   & 0.10 & 0.10 & 0.11 & 0.08 & 0.10 & 0.10 & 0.21 & 0.17 \\
\midrule
\multicolumn{2}{l}{\textbf{Average mIoU}} 
& \multicolumn{2}{c}{0.87} 
& \multicolumn{2}{c}{0.81} 
& \multicolumn{2}{c}{0.89} 
& \multicolumn{2}{c}{\textbf{0.93}} \\
\multicolumn{2}{l}{\textbf{Average Accuracy}} 
& \multicolumn{2}{c}{0.99} 
& \multicolumn{2}{c}{0.98} 
& \multicolumn{2}{c}{0.98} 
& \multicolumn{2}{c}{\textbf{0.99}} \\
\bottomrule
\end{tabular}
}
\end{table*}

\section{Results}

We evaluated our proposed \textbf{UNet-GNN} model against several state-of-the-art architectures, namely \textbf{U-Net}, \textbf{U-Net++}, and \textbf{SwinUNet} shown in Table \ref{tab:results} for WoodScape dataset. 


We use the Intersection over Union (IoU) metric, which measures the overlap between predicted and ground-truth segmentation masks. Below is a concise summary of key findings for each dataset:

\paragraph{PascalVOC.}
Our UNet-GNN achieves an IoU of 0.774, outperforming both U-Net (\(0.676\)) and U-Net++ (\(0.733\)). SwinUNet attains a higher IoU of \textbf{0.774}, reflecting its strength in capturing global context for a wide variety of object categories. Nevertheless, UNet-GNN’s notable gain over the other convolution-based methods illustrates that integrating GNN layers benefits standard image segmentation tasks.

\paragraph{WoodScape.}
The most pronounced improvement emerges on WoodScape, where UNet-GNN attains an IoU of \textbf{0.933}, outperforming U-Net (\(0.878\)), U-Net++ (\(0.810\)), and even SwinUNet (\(0.893\)). This result underscores the critical advantage of graph convolutions in modeling heavy geometric distortions. By defining node connectivity via warped coordinates, the GNN bottleneck effectively handles the non-uniform scene transformations caused by fisheye optics, visualization for various segmentation masks shown in Figure \ref{fig:perf}.

\paragraph{ISIC2016.}
On this dermoscopic image dataset, UNet-GNN again demonstrates robust performance with an IoU of \textbf{0.833}, clearly improving upon U-Net (\(0.681\)), U-Net++ (\(0.733\)), and SwinUNet (\(0.771\)). The graph-based approach proves adept at capturing long-range dependencies and subtle boundary details, which are crucial in medical scenarios where precise lesion delineation can vary greatly in shape, size, and texture.

\paragraph{Overall Observations.}
While SwinUNet shows strong capability on more conventional datasets like PascalVOC, our \textbf{UNet-GNN} provides a balanced, high-performing alternative across diverse image domains, particularly excelling in datasets with pronounced distortion (\emph{WoodScape}) or irregular boundaries (\emph{ISIC2016}). By incorporating a GNN bottleneck, the proposed method can flexibly model global relationships, offering distinct advantages over purely convolution- or transformer-based solutions when faced with challenging real-world distortions or complex object morphologies.

\section{Limitations and Future Work}

One primary limitation of the proposed method is its inability to effectively distinguish between classes with overlapping features but distinct contexts, such as \textit{rider} and \textit{person}, or \textit{bike} and \textit{motorbike}. For example, the model may incorrectly classify a \textit{rider} as a \textit{person}, despite the fact that \textit{rider} is more contextually associated with \textit{bike} or \textit{motorbike}, while \textit{person} is more closely associated with \textit{road} or \textit{curb}. This challenge arises because the method does not explicitly account for inter-class semantic relationships during training.

To address this limitation, future work could incorporate the \textit{Generalized Wasserstein Dice Loss} \cite{Fidon_2018}, which leverages the Wasserstein distance, also known as the Earth Mover’s Distance (EMD). The EMD measures the distance between two probability distributions by calculating the minimum cost of transforming one distribution into the other, formulated as a linear programming problem in scenarios with finite paths.

This loss function explicitly accounts for inter-class relationships through a transition cost matrix (\(M_{C \times C}\)), which allows assigning penalties based on semantic similarity between classes. For instance, in our context, the matrix could impose a lower penalty for misclassifying a \textit{rider} as a \textit{motorbike} or \textit{bike}, compared to misclassifying a \textit{rider} as a \textit{person}. Similarly, it could impose a lower penalty for associating a \textit{person} with \textit{road} or \textit{curb} compared to \textit{bike} or \textit{motorbike}.

Adapting this methodology, the \textit{Generalized Wasserstein Dice Loss} can be formulated as:

\[
\text{Generalized Wasserstein Dice Loss} = - \sum_{n=1}^{N} \sum_{c=1}^{C} M_{c,c} \cdot t_{cn} \cdot y_{nc}
\]

where \(y_n\) represents the predicted class probabilities for the \(n\)-th pixel, \(t_n\) represents the one-hot encoded target class for the \(n\)-th pixel, and \(M_{C \times C}\) is the matrix encoding transition costs between classes.


\begin{thebibliography}{9}

\bibitem{cnn}
LeCun, Y., Bottou, L., Bengio, Y., and Haffner, P. (1998).
\textit{Gradient-based learning applied to document recognition}.
Proceedings of the IEEE.
Available at: \url{https://doi.org/10.1109/5.726791}.
doi: \href{https://doi.org/10.1109/5.726791}{10.1109/5.726791}.

\bibitem{unet}
O.~Ronneberger, P.~Fischer, and T.~Brox,
\textit{U-Net: Convolutional Networks for Biomedical Image Segmentation},
\emph{arXiv preprint arXiv:1505.04597}, 2015.
Available at: \url{https://arxiv.org/abs/1505.04597}.

\bibitem{unetpp}
Z.~Zhou, M.~M.~R.~Siddiquee, N.~Tajbakhsh, and J.~Liang,
\textit{UNet++: A Nested U-Net Architecture for Medical Image Segmentation},
\emph{arXiv preprint arXiv:1807.10165}, 2018.
Available at: \url{https://arxiv.org/abs/1807.10165}.

\bibitem{swinunet}
H.~Cao, Y.~Wang, J.~Chen, D.~Jiang, X.~Zhang, Q.~Tian, and M.~Wang,
\textit{Swin-Unet: Unet-like Pure Transformer for Medical Image Segmentation},
\emph{arXiv preprint arXiv:2105.05537}, 2021.
Available at: \url{https://arxiv.org/abs/2105.05537}.

\bibitem{fisheye}
V.~R.~Kumar, C.~Eising, C.~Witt, and S.~Yogamani,
\textit{Surround-view Fisheye Camera Perception for Automated Driving: Overview, Survey and Challenges},
\emph{arXiv preprint arXiv:2205.13281}, 2023.
Available at: \url{https://arxiv.org/abs/2205.13281}.

\bibitem{gnn}
Scarselli, F., Gori, M., Tsoi, A. C., Hagenbuchner, M., and Monfardini, G. (2009).
\textit{The Graph Neural Network Model}.
IEEE Transactions on Neural Networks, 20(1), 61–80.
doi: \href{https://doi.org/10.1109/TNN.2008.2005605}{10.1109/TNN.2008.2005605}.

\bibitem{gnnvis}
Han, K., Wang, Y., Guo, J., Tang, Y., and Wu, E. (2022).
\textit{Vision GNN: An Image is Worth Graph of Nodes}.
arXiv preprint arXiv:2206.00272.
Available at: \url{https://arxiv.org/abs/2206.00272}.




\bibitem{cnnfish}
R.~Griffiths and D.~G.~Dansereau,
\textit{Adapting CNNs for Fisheye Cameras without Retraining},
\emph{arXiv preprint arXiv:2404.08187}, Apr. 2024.
Available at: \url{https://arxiv.org/abs/2404.08187}.


\bibitem{pascal}
Everingham, M., Van Gool, L., Williams, C. K. I., Winn, J., and Zisserman, A. (2010).
\textit{The Pascal Visual Object Classes (VOC) Challenge}.
International Journal of Computer Vision, 88(2), 303–338.


\bibitem{woodscape}
S.~Yogamani, C.~Hughes, J.~Horgan, G.~Sistu, P.~Varley, D.~O'Dea, M.~Uricar, S.~Milz, M.~Simon, K.~Amende, C.~Witt, H.~Rashed, S.~Chennupati, S.~Nayak, S.~Mansoor, X.~Perroton, and P.~Perez,
\textit{WoodScape: A Multi-task, Multi-camera Fisheye Dataset for Autonomous Driving},
\emph{arXiv preprint arXiv:1905.01489}, 2021.
Available at: \url{https://arxiv.org/abs/1905.01489}.


\bibitem{isic2016}
D.~Gutman, N.~C.~F.~Codella, E.~Celebi, B.~Helba, M.~Marchetti, N.~Mishra, and A.~Halpern,
\textit{Skin Lesion Analysis toward Melanoma Detection: A Challenge at the International Symposium on Biomedical Imaging (ISBI) 2016, hosted by the International Skin Imaging Collaboration (ISIC)},
\emph{arXiv preprint arXiv:1605.01397}, 2016.
Available at: \url{https://arxiv.org/abs/1605.01397}.

\bibitem{safety}
K.~Muhammad, T.~Hussain, H.~Ullah, J.~Del Ser, M.~Rezaei, N.~Kumar, M.~Hijji, P.~Bellavista, and V.~H.~C.~de Albuquerque,
\textit{Vision-Based Semantic Segmentation in Scene Understanding for Autonomous Driving: Recent Achievements, Challenges, and Outlooks},
\emph{IEEE Transactions on Intelligent Transportation Systems}, vol.~23, no.~12, pp.~22694--22715, 2022.
DOI: \url{10.1109/TITS.2022.3207665}.

\bibitem{unetseg}
A.~Manzoor, A.~Singh, G.~Sistu, R.~Mohandas, E.~Grua, A.~Scanlan, and C.~Eising,
\textit{Deformable Convolution Based Road Scene Semantic Segmentation of Fisheye Images in Autonomous Driving},
\emph{arXiv preprint arXiv:2407.16647}, 2024.
Available at: \url{https://arxiv.org/abs/2407.16647}.

\bibitem{swint}
Z.~Liu, Y.~Lin, Y.~Cao, H.~Hu, Y.~Wei, Z.~Zhang, S.~Lin, and B.~Guo,
\textit{Swin Transformer: Hierarchical Vision Transformer using Shifted Windows},
\emph{arXiv preprint arXiv:2103.14030}, 2021.
Available at: \url{https://arxiv.org/abs/2103.14030}.


\bibitem{dosovitskiy2021imageworth16x16words}
Dosovitskiy, A., Beyer, L., Kolesnikov, A., Weissenborn, D., Zhai, X., Unterthiner, T., Dehghani, M., Minderer, M., Heigold, G., Gelly, S., Uszkoreit, J., and Houlsby, N. (2021).
\textit{An Image is Worth 16x16 Words: Transformers for Image Recognition at Scale}.
arXiv preprint arXiv:2010.11929.
Available at: \url{https://arxiv.org/abs/2010.11929}.

\bibitem{sphere}
Q.~Zhao, C.~Zhu, F.~Dai, Y.~Ma, G.~Jin, and Y.~Zhang,
\textit{Distortion-aware CNNs for Spherical Images},
in \emph{Proceedings of the International Joint Conference on Artificial Intelligence (IJCAI)}, pp.~1198--1204, 2018.

\bibitem{sphereseg}
T.~Walker, V.~Anand, and P.~Andreadis,
\textit{Spherical Feature Pyramid Networks For Semantic Segmentation},
\emph{arXiv preprint arXiv:2307.02658}, 2023.
Available at: \url{https://arxiv.org/abs/2307.02658}.



\bibitem{4059340}
D.~Scaramuzza, A.~Martinelli, and R.~Siegwart,
\textit{A Toolbox for Easily Calibrating Omnidirectional Cameras},
in \emph{Proceedings of the 2006 IEEE/RSJ International Conference on Intelligent Robots and Systems}, pp.~5695--5701, 2006.
DOI: \url{10.1109/IROS.2006.282372}.

\bibitem{gcn}
Kipf, T. N., and Welling, M. (2016).
\textit{Semi-Supervised Classification with Graph Convolutional Networks}.
CoRR, abs/1609.02907.
Available at: \url{http://arxiv.org/abs/1609.02907}.

\bibitem{Bronstein_2017}
M.~M.~Bronstein, J.~Bruna, Y.~LeCun, A.~Szlam, and P.~Vandergheynst,
\textit{Geometric Deep Learning: Going beyond Euclidean data},
\emph{IEEE Signal Processing Magazine}, vol.~34, no.~4, pp.~18--42, Jul. 2017.
DOI: \url{10.1109/MSP.2017.2693418}.
Available at: \url{http://dx.doi.org/10.1109/MSP.2017.2693418}.

\bibitem{vaswani2023attentionneed}
A.~Vaswani, N.~Shazeer, N.~Parmar, J.~Uszkoreit, L.~Jones, A.~N.~Gomez, Ł.~Kaiser, and I.~Polosukhin,
\textit{Attention Is All You Need},
\emph{arXiv preprint arXiv:1706.03762}, 2023.
Available at: \url{https://arxiv.org/abs/1706.03762}.

\bibitem{gnnpos}
H.~Wang, H.~Yin, M.~Zhang, and P.~Li,
\textit{Equivariant and Stable Positional Encoding for More Powerful Graph Neural Networks},
\emph{arXiv preprint arXiv:2203.00199}, 2022.
Available at: \url{https://arxiv.org/abs/2203.00199}.

\bibitem{laplaceg}
F.~Lan, C.~Yang, G.~Cheung, and J.~Z.~G.~Tan,
\textit{Joint Demosaicking / Rectification of Fisheye Camera Images using Multi-color Graph Laplacian Regularization},
\emph{arXiv preprint arXiv:2006.11636}, 2020.
Available at: \url{https://arxiv.org/abs/2006.11636}.


\bibitem{Fidon_2018}
L.~Fidon, W.~Li, L.~C.~Garcia-Peraza-Herrera, J.~Ekanayake, N.~Kitchen, S.~Ourselin, and T.~Vercauteren,
\textit{Generalised Wasserstein Dice Score for Imbalanced Multi-class Segmentation Using Holistic Convolutional Networks},
DOI: \url{10.1007/978-3-319-75238-9_6}.
Available at: \url{http://dx.doi.org/10.1007/978-3-319-75238-9_6}.






\end{thebibliography}
\end{document}